\documentclass[10pt,twocolumn,letterpaper]{article}

\usepackage[final]{cvpr}      
\usepackage{algorithm}
\usepackage{algorithmic}
\usepackage{multirow}
\usepackage[T1]{fontenc}
\definecolor{cvprblue}{rgb}{0.21,0.49,0.74}
\usepackage[pagebackref,breaklinks,colorlinks,allcolors=cvprblue]{hyperref}

\def\paperID{7143} 
\def\confName{CVPR}
\def\confYear{2025}

\title{GUSLO: General and Unified Structured Light Optimization}

\author{Tinglei Wan\\
Harbin Institute of Technology\\
\and
Tonghua Su\\
Harbin Institute of Technology\\
\and
Zhongjie Wang\\
Harbin Institute of Technology\\
}

\begin{document}
\maketitle
\begin{abstract}
Structured light (SL) 3D reconstruction captures the precise surface shape of objects, providing high-accuracy 3D data essential for industrial inspection and cultural heritage digitization. However, existing methods suffer from two key limitations: reliance on scene-specific calibration with manual parameter tuning, and optimization frameworks tailored to specific SL patterns, limiting their generalizability across varied scenarios.
We propose \textbf{G}eneral and \textbf{U}nified \textbf{S}tructured \textbf{L}ight \textbf{O}ptimization (\textbf{GUSLO}), a novel framework addressing these issues through two coordinated innovations: (1) single-shot calibration via 2D triangulation-based interpolation that converts sparse matches into dense correspondence fields, and (2) artifact-aware photometric adaptation via explicit transfer functions, balancing generalization and color fidelity.
We conduct diverse experiments covering binary, speckle, and color-coded settings. Results show that GUSLO consistently improves accuracy and cross-encoding robustness over conventional methods in challenging industrial and cultural scenarios.
\end{abstract}
    
\section{Introduction}

Structured light (SL) 3D reconstruction has become a vital technique in industrial and cultural heritage applications due to its high precision and non-contact nature. In industry, SL enables rapid reverse engineering~\cite{qian2019high,bak2003rapid} and real-time quality control~\cite{qian2021high,sansoni2009state}. In cultural heritage preservation, SL is widely adopted for digitizing and monitoring fragile artifacts, such as historical clothing~\cite{montusiewicz2021structured,ding2024digital}, architectural surfaces under climatic change~\cite{holl2021structured}, and detailed edge visualization in 3D point clouds~\cite{yamada2024high}. It also supports deformation analysis of organic materials~\cite{stelzner2022stabilisation}, contributing to long-term conservation workflows.

\begin{figure}[t]
\centering
  \includegraphics[width=\linewidth]{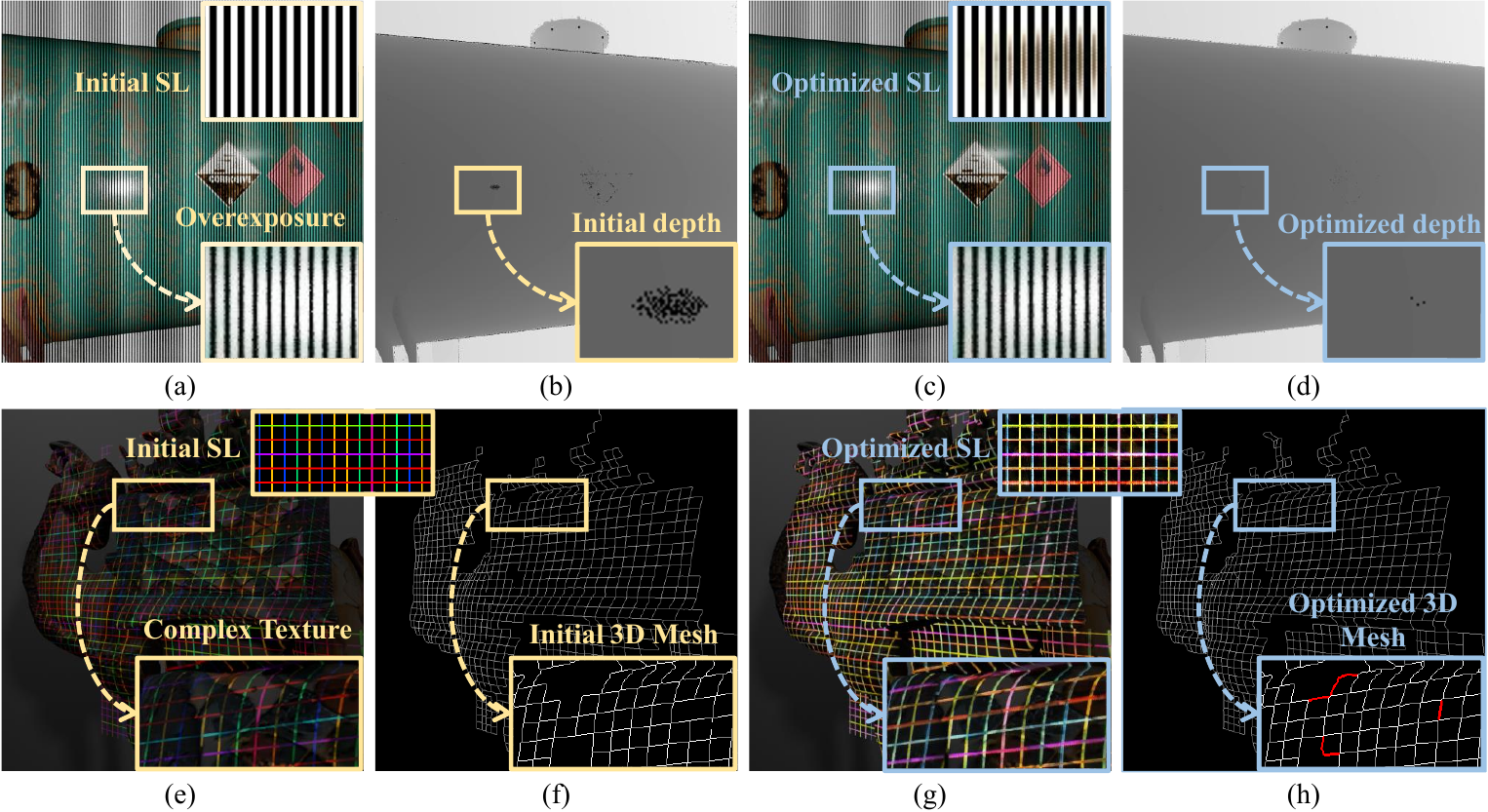}
    \caption{Robust 3D reconstruction with our framework. (a–b) Overexposure disrupts pattern coding, leading to missing geometry. (c–d) Our projection compensates illumination loss, enabling complete reconstruction. (e–f) Surface texture degrades color-coded decoding, causing partial mesh loss. (g–h) Texture-aware chromatic optimization restores decoding and improves reconstruction completeness (highlighted in red).}
    \label{exp_rendering}
\end{figure}

A core challenge in SL lies in optimizing projection patterns under complex lighting and surface textures. Traditional methods focus on two non-adaptive strategies: post-processing camera measurements (e.g., denoising~\cite{dodda2023denoising}, phase correction~\cite{wang2024use}) and heuristic pattern design (e.g., composite~\cite{nguyen2022single}, frequency~\cite{li2022deep}, or color multiplexing~\cite{fu2024deep}). These hand-crafted approaches fail to adapt to photometric variations, limiting robustness in uncontrolled environments, as illustrated in Fig.~\ref{exp_rendering}.

Projection compensation methods attempt to address this by modeling the projection process to adaptively refine input patterns~\cite{shih2020enhancement,sugimoto2021directionally}. Though effective in reducing ambient and reflectance-induced color shifts~\cite{fujii2005projector,li2023physics}, they face two key limitations: (1) they rely on extensive calibration to map surface reflectance to the projector plane, which restricts their applicability to simple geometries; and (2) they require multiple projections per scene, making them impractical for real-time or general-use deployment.

To overcome these challenges, we propose a general and unified SL optimization (GUSLO) framework (Fig.~\ref{System_framework}) that achieves joint photometric calibration and SL pattern optimization using a single projection. Our approach is based on two key components: (1) continuous surface parameterization via barycentric interpolation over triangulated discrete matches, and (2) physics-aware photometric transfer through enhanced thin-plate spline (TPS) warping with adaptive color mapping.

First, the projector-camera global matching process employs planar triangulation with barycentric interpolation to parameterize 3D surfaces into a continuous UV mapping space. This geometric foundation facilitates single-shot calibration via projection-plane conformal mapping. 
Second, the SL projection compensation process employs an enhanced thin-plate spline (TPS) operator to achieve adaptive artifact-aware photometric transfer. This framework enables dynamic color mapping while preserving gradient coherence across illumination conditions.
Our work introduces fundamental advancements in structured light optimization through three key innovations:
\begin{itemize}
\item \textbf{General and Unified SL Optimization Architecture}:
We propose \textbf{GUSLO}, a novel structured light optimization framework that jointly performs geometric calibration and photometric compensation under a single-projection setup. Extensive experiments on binary, speckle, and color-coded SL patterns demonstrate GUSLO’s superior cross-encoding generality and significant accuracy improvements across diverse industrial and cultural heritage scenarios.
\item \textbf{Single-Projection Geometric-Photometric SL Optimization}:
To the best of our knowledge, GUSLO is the first SL optimization system that enables \emph{simultaneous geometric and photometric calibration using a single structured pattern}. Our design resolves the long-standing adaptability-accuracy trade-off in SL systems by integrating continuous projective-plane parameterization with dynamic, artifact-suppressed color correction.
\item \textbf{Open and Reproducible Benchmarking System}:
We establish an open-source simulation and evaluation platform for SL research, featuring a physically-grounded calibration setup, configurable virtual scenes, and standardized protocols. Built upon Blender and parametric scripting, the system enables automated dataset generation and reproducible benchmarking for future research.
\end{itemize}
\begin{figure*}[t]
  \centering
    \includegraphics[width=0.95\linewidth]{figures/pipeline.pdf}
  \caption{Model details of our GUSLO. Stage-I filters erroneous codes based on relative positions in the camera plane, then applies Delaunay triangulation and barycentric interpolation to extend discrete matches into continuous global correspondences between camera and projector pixels. In Stage-II, decoding artifacts are corrected via the DeArtifact Module, implemented using our WC-TPS method trained with ShadingNet data. The refined results are then processed by PCNet for photometric compensation, yielding optimized structured light patterns for high-precision 3D reconstruction.}
  \label{System_framework}
\end{figure*}

\section{Related Work}
Our proposed unified framework for SL optimization consists of two main components: projector-camera global matching and projection compensation. We first introduce related work in these two areas. Additionally, we introduce various recent attempts by researchers to generate SL patterns that are better adapted to the capturing scene.
\subsection{Projector-Camera Global Matching}

Matching methods can be broadly categorized by surface continuity. For planar surfaces, Raskar et al.~\cite{raskar2002low} introduced a 3$\times$3 response matrix, while Huang et al.~\cite{huang2018single} extended it to curved surfaces via TPS and affine transformations, though requiring scene-specific training data.

For non-continuous surfaces, Gupta et al.~\cite{gupta2007gray} proposed composite Gray code patterns, and Pages et al.~\cite{1241585} used sinusoidal patterns for phase computation, both requiring multiple projections and incurring time overhead.

Hardware-assisted methods, such as coaxial optics~\cite{fujii2005projector} or RGBD cameras with differentiable rendering~\cite{park2022projector}, can accelerate matching but depend on specialized setups, limiting practical deployment.

\subsection{Projector Compensation} 

Traditional compensation approaches modify projector input based on environment lighting. For example, Raskar et al.~\cite{raskar2002low} derived a color-space compensation function via input-output mapping, but such methods struggle with complex reflectance and often rely on discrete Gray code mappings.

Deep learning approaches improve adaptability. Huang et al.~\cite{huang2021end} separated projection and surface components, and Park et al.~\cite{park2022projector} used differentiable rendering to optimize input images. However, both require scene-specific retraining or additional hardware, limiting real-time and general-purpose SL applications.

\subsection{Structured Light Projection Optimization}

Recent work focuses on tailoring SL patterns to capture conditions. Xu et al.~\cite{xu2023unified} used LED arrays and LCD masks to learn optimal encoding sets, combining results across LEDs. Dong et al.~\cite{dong2023adaptive} proposed ambient-light-adaptive color SL using MAP-based color detection. Jia et al.~\cite{jia2024adaptivestereo} designed depth-adaptive speckle SL by adjusting gray level, density, and size across sub-regions.

While effective for specific inputs, these methods lack generality. Xu's method targets LED-based SL, Dong's is unsuitable for grayscale, and Jia's focuses solely on speckle. As such, none of them generalize well across diverse SL types or scenes.

\section{Methodology}
Our GUSLO framework (Fig.~\ref{System_framework}) is structured around two core processes: (1) Delaunay-barycentric interpolation to generate dense geometric mappings from sparse correspondences, and (2) a physics-aware compensation scheme that integrates thin-plate spline warping with photometric modeling to address projector nonlinearities and ambient illumination.
The optimization flow first builds continuous surface parameterization via the \textbf{Projector-Camera Global Matching} stage, then refines SL inputs through the \textbf{Structured Light Projection Compensation} stage, enabling cross-scene and cross-encoding optimization from a single projection. To support reproducible evaluation, our \textbf{Open Structured Light Benchmarking System} offers hardware-agnostic validation through virtual co-calibration and automated dataset generation.

\subsection{Projector-camera Global Matching}\label{s3.1}
\textbf{Matching Pattern Design.}
Accurate pixel localization between projector and camera is essential for SL pattern optimization.
We designed SL images using De Bruijn \cite{kawasaki2008dynamic} sequences for this purpose. In our method, the horizontal alphabet \( A_{hor} = \{1, 3, 5, 7\} \) corresponds to red, lime, cyan, and purple, while the vertical alphabet \( A_{ver} = \{2, 4, 6, 8\} \) represents yellow, green, blue, and magenta. With a window length of 3, no sequence of three consecutive lines repeats, allowing unique identification of stripe intersections. To increase the number of matching feature points, orange-red dots are placed at the center of the grid.

To avoid interference with stripe detection, the pattern background color is selected within a moderate RGB range, ensuring color separation from grid elements without causing clipping.

\textbf{Discrete Matching and Error Code Filtering.}
Project the calibration pattern (shown in Fig. \ref{System_framework}) onto the object surface, where the camera captures the deformed pattern (see Camera Captures in Fig. \ref{System_framework}, top-left).
Then use De Bruijn coding to match feature points between the projected and captured patterns. This results in discrete sets of projector-camera matching points, \( P = \bigl\{ p_i = (x_{p}^{i}, y_{p}^{i}),\ i = 1, 2, \dots, n \bigr\} \) and \( C = \bigl\{ c_i = (x_{c}^{i}, y_{c}^{i}),\ i = 1, 2, \dots, n \bigr\} \). The matching relationship between these points can be described as a bijection 
\begin{equation}\label{eq1}
M\text{=}\left\{ \left(p_i,c_i\right)| p_i\in P,c_i\in C,i\text{=1,2,…,}n \right\}
\end{equation}
from the projector points \( P \) to the camera points \( C \).

During the matching process, De Bruijn code losses in some regions can lead to decoding errors in the grid's central feature points.
These errors are filtered out to avoid propagating mismatches.

We compute the horizontal and vertical relative position matrices, \( X_{map}^{p} \) and \( Y_{map}^{p} \) on the projector image plane, and \( X_{map}^{c} \) and \( Y_{map}^{c} \) on the camera image plane, by comparing the relative positions of the feature points \( (x_{p}^{i}, y_{p}^{i}) \) and \( (x_{c}^{i}, y_{c}^{i}) \) with their surrounding eight feature points. Eq. \ref{eq12} presents the XOR of these matrices, indicating the similarity between the two sets of position encodings.
\begin{equation}\label{eq12}
V_x=X_{map}^{p}\oplus X_{map}^{c} ,\quad V_y=Y_{map}^{p}\oplus Y_{map}^{c}.
\end{equation}

Finally, by performing row summation on \(V_x\) and \(V_y\), we can obtain the voting scores for error codes in the \(x\) and \(y\) analyses, \(S_x\) and \(S_y\), for each point.
Let the elements in \( S_x \) and \( S_y \) be \( S_{x}^{i} \) and \( S_{y}^{i} \), respectively. The set of filtered point indices, denoted by \textit{indices}, is defined as:
\begin{equation}\label{eq14}
\textit{indices}=\{i\mid i\in \{1,2,...,n\}\,\,\text{and\,\,}\left( S_{x}^{i}\ge k\,\,\text{or\,\,}S_{y}^{i}\ge k \right) \}.
\end{equation}
The value of \( k \) is related to the number of points \( n \) involved in the error code check. Generally, \( k \) is set as \( \text{Floor}(n/2) \), where \( \text{Floor}(\cdot) \) denotes the floor function.

\textbf{Global Matching.}
We employ Delaunay triangulation to convert sparse matches into continuous mappings over the projector image plane.
This extends \( M \) to every pixel on the projector image plane by triangulating the point set \( P \), forming a continuous triangular mesh where each vertex \( p_1, p_2, p_3 \) corresponds to camera image coordinates \( c_1, c_2, c_3 \).

For any point \( p = (x_p, y_p) \) within a triangular patch, its barycentric coordinates calculated using Eq. \ref{eq4} to \ref{eq7}.
\begin{equation}\label{eq4}
\det T=\left( y_{p}^{2}-y_{p}^{3} \right) \left( x_{p}^{1}-x_{p}^{3} \right) +\left( x_{p}^{3}-x_{p}^{2} \right) \left( y_{p}^{1}-y_{p}^{3} \right),
\end{equation}
\begin{equation}\label{eq5}
L_1=\frac{\left( y_{p}^{2}-y_{p}^{3} \right) \left( x_p-x_{p}^{3} \right) +\left( x_{p}^{3}-x_{p}^{2} \right) \left( y_p-y_{p}^{3} \right)}{\det T},
\end{equation}
\begin{equation}\label{eq6}
L_2=\frac{\left( y_{p}^{3}-y_{p}^{1} \right) \left( x_p-x_{p}^{3} \right) +\left( x_{p}^{1}-x_{p}^{3} \right) \left( y_p-y_{p}^{3} \right)}{\det T},
\end{equation}
\begin{equation}\label{eq7}
L_3=1-L_1-L_2.
\end{equation}

If \( p \) lies within the triangle formed by \( p_1, p_2, p_3 \), then \( L_1\ge 0\), \(L_2\ge 0\), and \(L_3\ge 0 \). The corresponding texture coordinates \( c(x_c, y_c) \) can be obtained through barycentric interpolation.
\begin{equation}\label{eq8}
\begin{bmatrix}
x_c \\ 
y_c
\end{bmatrix}
= 
L_1 \begin{bmatrix}
x_c^1 \\
y_c^1
\end{bmatrix}
+
L_2 \begin{bmatrix}
x_c^2 \\
y_c^2
\end{bmatrix}
+
L_3 \begin{bmatrix}
x_c^3 \\
y_c^3
\end{bmatrix}
\end{equation}

By traversing each triangular patch on the projector image plane and performing barycentric interpolation for the points inside, corresponding camera coordinates are calculated, establishing a continuous global mapping between the projector and the camera.

\subsection{Structured Light Projection Compensation}
\textbf{Compensation Process Modeling.}
To achieve adaptive adjustment of the projection input based on lighting and object reflectance, we need to model the projection process, which is represented by Eq. \ref{eq19}.
\begin{equation}\label{eq19}
\boldsymbol{x}^* = \mathcal{F}^{\dag} \left( \boldsymbol{x}; \boldsymbol{\tilde{s}} \right),
\end{equation}
where \(\boldsymbol{\tilde{s}}\) is the surface image of the object under global illumination, \(\boldsymbol{x}^*\) is the SL projection pattern, and \(\boldsymbol{x}\) is the projection pattern captured by the camera. To ensure that the projection pattern is unaffected by ambient light or object surface texture, we set the initial pattern as \(\boldsymbol{x}_0\) and the ideal captured image as \(\boldsymbol{x}=\boldsymbol{x}_0\). The optimized projection pattern is obtained using:
\begin{equation}\label{eq20}
\widehat{\boldsymbol{x}}=\mathcal{F}^{\dag}\left(\boldsymbol{x}_0;\boldsymbol{\tilde{s}}\right).
\end{equation}

Optimizing the SL pattern requires solving the inverse projection function \( \mathcal{F}^{\dag} \). To generalize this solution, the photometric compensation network (PCNet), is trained to model \( \mathcal{F}^{\dag} \) using globally matched projector-camera image pairs \( \left( \boldsymbol{\tilde{x};x} \right) \) and corresponding surface images \( \boldsymbol{\tilde{s}} \). PCNet consists of a siamese encoder and a decoder, as shown in Fig. \ref{System_framework}. The encoders share weights, and by subtracting the features learned from \( \boldsymbol{\tilde{s}} \) from those learned from \( \boldsymbol{\tilde{x}} \), we can separate the varying parts of ambient light, object surface texture, and reflections from the overall photometric model. This allows the network to focus on the photometric effects caused by the projection, giving the photometric compensation network generalization capabilities.

However, many pixels in this process deviate from physical laws. For example, if blue light is projected onto a surface that absorbs it, adjusting the blue channel of \( \widehat{\boldsymbol{x}} \) via Eq.~\ref{eq20} may not reproduce the effect of \( \boldsymbol{x}_0 \). This can cause brightness or chromaticity clipping, resulting in image artifacts. To address this, we introduce an additional module that minimally refines the optimization objective to achieve physically plausible results.

\textbf{DeArtifact Module.}
We develop a Weighted-Constrained Thin-Plate Spline (WC-TPS) method to regulate PCNet's input within physically valid brightness clipping bounds. WC-TPS models compensation as a 3D TPS deformation governed by Eq.~\ref{eq21}, enforcing spatial smoothness via spline regularization and clipping constraints through boundary conditions. This dual mechanism redistributes compensation gradients across non-clipping regions while preserving sub-pixel optical continuity, translating TPS smoothness into photometrically constrained adjustments.

\begin{equation}\label{eq21}
\boldsymbol{x}^*=f\left(\boldsymbol{x}^{\prime}\right)=\sum_{i=0}^{N-1}{\omega _i}\phi \left(\parallel \boldsymbol{x}^{\prime}-\boldsymbol{x}_{i}^{\prime}\parallel \right)+\mathbf{a}^T\boldsymbol{x}^{\prime}+b,
\end{equation}
where \(\boldsymbol{x}^*\) is the compensation result, \(\boldsymbol{x}^{\prime}\) is the captured image, and \(\omega_i\), \(\mathbf{a}\), and \(b\) are parameters determined by solving a linear system of equations.
To compute the TPS projection compensation function, \(N \left(N\geq4\right)\) sets of projection compensation image pairs are needed. During global matching, a real projection compensation data pair is obtained, while the remaining \(N-1\) simulated pairs need to be generated using synthetic data. To ensure that the generation process aligns with the current scenario, we know from \ref{eq20} that the projection process is the inverse of the compensation process. Thus, we train a projection network (ShadingNet) suitable for the current scenario using a network architecture similar to PCNet, with the input and output positions swapped during training. By using uniformly distributed solid-color images in the color space as projection inputs, we can generate the corresponding projection results for the current scenario via ShadingNet, providing reliable synthetic data for TPS.
During the parameter computation process, the real data pair \( \left( \boldsymbol{x}_{\boldsymbol{real}}^{*}, \boldsymbol{x}_{\boldsymbol{real}}^{\prime} \right) \) has a higher weight:
\begin{equation}\label{eq26}
\mathbf{w}=\left( \alpha w_1,\alpha w_2,...,\left(1-\alpha\right) w_N\right)^T\quad \text{where\quad } \alpha = 0.35.
\end{equation}

To minimize clipping, we adopt the optimization method proposed by \cite{grundhofer2015robust} to smoothly adjust the input brightness.
\begin{equation}\label{eq27}
\boldsymbol{x}_{0_{adapt}}=\boldsymbol{s*x}_0,
\end{equation}
First, we define the error functions that will be minimized in the optimization process. The error function includes the saturation error \( err_{\text{sat}} \), which prevents color clipping, the gradient variation error \( err_{\text{grad}} \), which ensures smooth image adjustments, and the intensity error \( err_{\text{int}} \), which maintains overall brightness.

Next, the variable brightness scaling value \( S \) is adjusted non-linearly for each pixel. The optimal \( S \) is found by minimizing the following objective:
\begin{equation}\label{eq28}
S_{opt} = \mathop {\mathrm {argmin\kern 0pt}} _{S} err_{opt}(S).
\end{equation}
The total error function is:
\begin{equation}\label{eq29}
\begin{aligned}
err_{opt}(S) = &\omega_{\text{sat}} \cdot err_{\text{sat}}(S) + \omega_{\text{grad}} \cdot err_{\text{grad}}(S) \\
& + \omega_{\text{int}} \cdot err_{\text{int}}(S).
\end{aligned}
\end{equation}
Here, \( \omega_{\text{sat}} \), \( \omega_{\text{grad}} \), and \( \omega_{\text{int}} \) are the weights for each error term.

To mitigate potential inaccuracies in TPS-based photometric compensation, we introduce a photometric inspector that uses PCNet-generated solid-color projections spanning the RGB space. Since TPS operates pixel-wise, it may produce color discontinuities, detected via gradient magnitude. To avoid misclassifying object texture edges as artifacts, regional structural similarity is also evaluated. Pixels failing either criterion are masked out, bypassing DeArtifact and reverting to direct PCNet optimization.
\begin{equation}
\label{eq33}
M(p) = 
\begin{cases} 
1 & \text{if } G_p \geq \tau_g \text{ or } \text{SSIM}(I_{A(p)},R_{A(p)}) \leq \tau_s\\
0 & \text{otherwise}
\end{cases}
\end{equation}

Here, $G_p$ represents the color gradient magnitude at pixel~$p$, and $\text{SSIM}$ computes the structural similarity between the local image region~$A(p)$ and its reference counterpart. The binary mask~$M(p)$ is determined by comparing these features against two optimized thresholds: $\tau_g$ for gradient sensitivity and $\tau_s$ for structural dissimilarity, where $\tau_g$ controls texture edge detection while $\tau_s$ governs allowable photometric deviations.

\subsection{Open Structured Light Benchmarking System}
We present an open-source benchmarking system for structured light research, addressing three key limitations of conventional methods: (1) prolonged calibration of complex hardware setups, (2) scene-specific constraints in real-world data collection, and (3) lack of reliable ground-truth for 3D reconstruction.

Our Blender-based \cite{blender2025} virtual calibration system emulates projector-camera setups for geometry-consistent simulation. It supports flexible radiometric configurations (e.g., BRDFs, dynamic lighting) and geometric distortion modeling (e.g., specularities, complex textures), with automated multi-parameter optimization via scripting to generate large-scale radiometric and geometric datasets.

The system defines standardized SL evaluation protocols, including programmable encoding tests, intrinsic/extrinsic calibration using virtual targets, and multi-level quantitative metrics for texture fidelity and 3D accuracy. By decoupling evaluation from physical hardware, it enables reproducible testing and cross-platform comparison. We will release the full codebase and preset scenes to support standardized, verifiable SL research.

\section{Experiments}
We adopt a two-stage training strategy to enhance the generalization of the photometric compensation network. PCNet and ShadingNet are first pre-trained on 150 synthetically generated datasets via our procedural pipeline, then fine-tuned on 40 real-world measurements under diverse illumination, combining 20 samples from \cite{huang2019end} and 20 captured by our multi-projector array.

Each dataset contains 700 projection images per object surface, with 500 for training and 200 for testing. During training, object surface images and their corresponding projections are randomly paired to disentangle global illumination from surface reflections. The calibration pattern’s background color is set to \( \mathbf{v} = (0.25, 0.25, 0.25) \). For Eq.~\ref{eq33}, we use \( \tau_g = 20 \) and \( \tau_s = 0.8 \).

We validate our global matching algorithm against Gray code and deep learning baselines, focusing on continuous projector-camera mapping. Then, we evaluate optimization performance across multiple SL patterns, establishing the first unified optimization framework across SL types. Decoding accuracy is used as the evaluation metric. All experiments are conducted using a Hikvision MV-CU013-A0GC camera (1280×1024) and a Sony VPL-EX570 projector (1024×768).

\subsection{Comparison of Global Matching Methods}
We compare our global matching method (GMM) with traditional Gray code (GC) and deep learning-based WarpingNet~\cite{huang2021end}. GC requires 42 patterns to encode each projector pixel, but it produces only discrete correspondences and fails to establish continuous projector-camera mappings. This results in decoding artifacts, especially at codeword transitions and in regions with resolution mismatch.

\begin{figure}[ht]
  \centering  \includegraphics[width=0.95\linewidth]{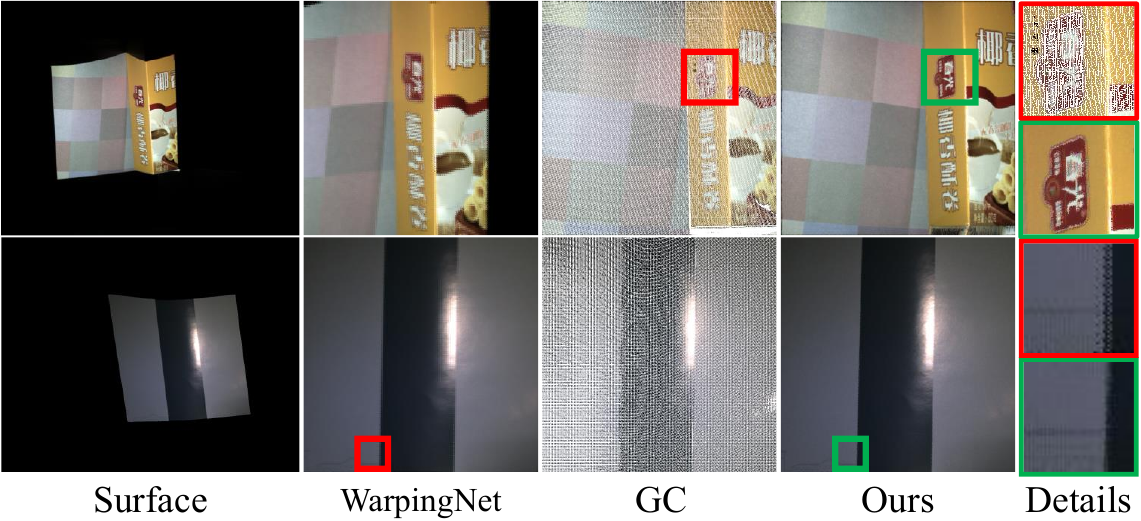}
  \caption{Comparison with the method of global encoding using Gray code (GC) and WarpingNet.}
  \label{exp1.1}
\end{figure}

Our method addresses these limitations using a single projection, applying De Bruijn-based sparse matching, error-code filtering, and Delaunay-based interpolation to produce smooth, continuous mappings. As shown in Fig.~\ref{exp1.1}, our results demonstrate better texture continuity and fewer artifacts.

We also compare with WarpingNet, which is trained on 500 projected images. While effective for single-object scenes, its performance drops significantly on multi-object or textured surfaces. Fig.~\ref{exp1.1} shows that WarpingNet suffers from interpolation blur and cross-object mismatch, whereas GMM preserves finer details.

To ensure a fair comparison, we apply linear interpolation and our GMM to the 42-projection GC outputs for continuous mapping. Table~\ref{tab:GMM} presents the results: our GMM reduces error rate by 60.31\% compared to GC, and achieves only 3.67\% higher error than GC+linear. The addition of our EC filter further improves accuracy by 14.01\%.

\begin{table}[htbp]
\centering
\caption{Our method (GMM) achieves reliable matching results using only a single image.}
\label{tab:GMM}
\setlength{\tabcolsep}{1mm}
{\fontsize{9}{10}\selectfont
\begin{tabular}{|c|c|c|c|c|}
\hline
\multirow{2}{*}{\textbf{Method}} & \multicolumn{2}{c|}{\textbf{Continuous Surface}} & \multicolumn{2}{c|}{\textbf{Discontinuous Surface}} \\ \cline{2-5}
& SSIM$\uparrow$ & Err. (\%)$\downarrow$ & SSIM$\uparrow$ & Err. (\%)$\downarrow$ \\ \hline
WarpingNet & 0.813 & 4.61  & N/A   & N/A   \\ \hline
GC         & 0.361 & 5.42  & 0.362 & 5.63  \\ \hline
GMM        & 0.912 & 2.15  & 0.889 & 3.38  \\ \hline
GC+Linear  & 0.935 & 2.09  & 0.927 & 3.35  \\ \hline
GC+GMM     & \textbf{0.957} & \textbf{2.07}  & \textbf{0.952} & \textbf{3.31} \\ \hline
\end{tabular}
}
\end{table}

\begin{figure*}[th]
    \centering  \includegraphics[width=\textwidth]{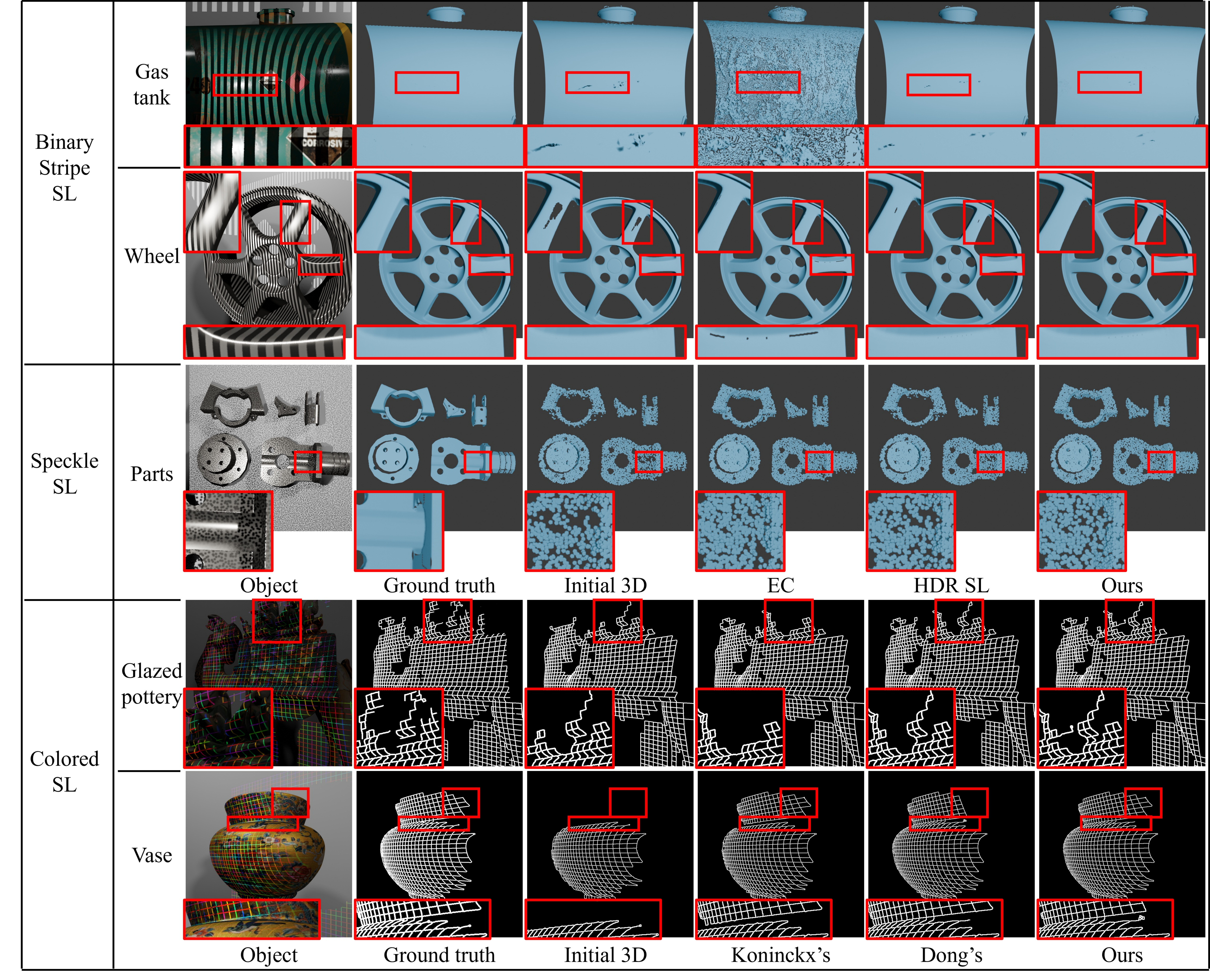}
    \caption{Visual comparison of optimized structured light patterns under different conditions: darkroom (rows 1 \& 4), multi-lighting (rows 2, 3, 5), Lambertian (rows 2 \& 3), non-Lambertian (rows 1, 4, 5), textured (rows 1, 4, 5), and textureless surfaces (rows 2 \& 3). Our optimized patterns yield more complete and accurate geometry across all scenarios.}
    \label{exp_all_q}
\end{figure*}

\subsection{Optimization Effects of Different Structured Light Patterns}
We evaluate our method on three representative SL patterns—binary stripes (e.g., Gray code), speckle, and color-coded—under diverse lighting, texture, and reflectance conditions (see Fig.~\ref{exp_all_q}). Captured images are mapped into projector space to enable pixel-wise decoding error analysis before and after optimization. Additional results, including variations in exposure and pattern parameters, are provided in the supplementary material.

\subsubsection{Binary Stripe Structured Light Pattern}
Gray code is widely used in binary encoding. We compare our method with HDR SL~\cite{fu2023high}, which adjusts light intensity to mitigate overexposure, and Exposure Correction (EC)~\cite{Zou_2024_CVPR}, which post-processes captured images.

To ensure fairness, all methods use the same projection intensity and decoding pipeline. HDR SL is grayscale-only and lacks spatial adaptability; EC often overcompensates for dark textures, mistaking them for underexposure.
In contrast, our method adapts to both surface texture and illumination, achieving a 3.05\% average absolute error reduction and a 6.34\% relative improvement over HDR SL. It also outperforms EC by at least 5.98\% across all eight scenarios. Fig.~\ref{exp_all_q} (rows 1–2) shows more complete depth reconstruction.

\subsubsection{Speckle Structured Light Pattern}
Speckle SL is widely used in stereo-based matching but is sensitive to reflectance and exposure variations. We compare our method with HDR SL and EC under consistent speckle generation and decoding parameters, isolating the impact of photometric optimization.

As shown in Fig.~\ref{exp_all_q}, EC struggles with highlight and low-contrast regions, while our method reduces artifacts effectively. It yields an 8.70\% average absolute error reduction, with 13.84\% and 54.41\% relative improvements over HDR SL and EC, respectively.
We further evaluate performance under varying exposure times and speckle densities/sizes (see the supplementary material). Our method consistently delivers higher decoding accuracy. 

\subsubsection{Colored Structured Light Pattern}
Color-coded SL allows high-density single-shot reconstruction. We optimize De Bruijn stripe patterns and compare with Koninckx et al.~\cite{koninckx2005scene}, which calibrates projector-camera response curves, and Dong et al.~\cite{dong2023adaptive}, which performs unsupervised color detection.

For fairness, we pre-calibrate response curves for Koninckx’s method and use the same De Bruijn patterns and resolution across methods. Koninckx achieves only localized correction in overexposed regions ($\le$12.7\%), with a 5.55\% average error reduction. Dong's method achieves 14.16\% via uniform color segmentation but sacrifices spatial discriminability.
Our method integrates physical modeling with adaptive chromatic mapping, achieving a 29.39\% average error reduction and artifact-free decoding (see Fig.~\ref{exp_all_q}).

\section{Ablation Study}
To systematically validate our proposed technical components, we conduct ablation studies focusing on two critical modules: (1) the global matching module with error code filtering, and (2) the DeArtifact module for SL projection compensation. Both components address fundamental challenges in structured light reconstruction through distinct technical approaches.

\begin{figure}[ht]
  \centering
  \includegraphics[width=0.8\linewidth]{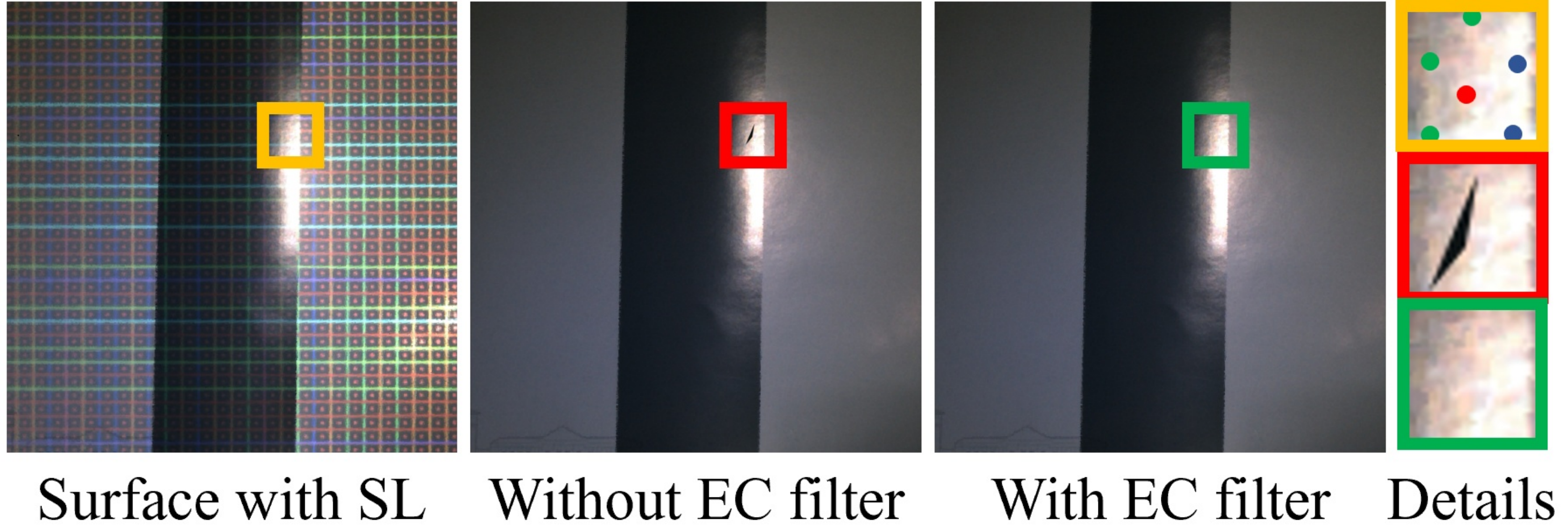}
  \caption{Global matching results with and without the error code (EC) filter. The fourth column is an enlarged view of the erroneous matches, with red dots indicating incorrect matches, green dots indicating correct matches, and blue dots indicating unmatched points.}
  \label{ablation_1}
\end{figure}

\textbf{Global Matching with Error Code Filtering:} 
In the projector-camera global matching module, we employ error code filtering to reduce feature point matching errors during global matching. To demonstrate the effectiveness of this module, we conducted an ablation experiment by removing the error code filter and performing triangulation and global mapping on the original matching pairs. The comparative results are shown in Fig. \ref{ablation_1}. 

When there is continuous loss in the surrounding De Bruijn encoding, the orange-red dots may experience decoding errors, leading to incorrect placement. This can result in erroneous triangle color mapping, as shown in the figure. After filtering out these errors, the region is correctly matched through re-triangulation and color mapping within the triangles. Quantitative evaluation of the EC filter's effectiveness, presented in Table \ref{tab:ablation_ec}.

\begin{table}[htbp]
\centering
\caption{The EC filter significantly reduces the maximum decoding error and improves optimization stability}
\label{tab:ablation_ec}
\setlength{\tabcolsep}{0.8mm}
{\fontsize{9}{10}\selectfont
\begin{tabular}{|c|c|c|c|c|c|}
\hline
\textbf{Model} & SSIM$\uparrow$ & PSNR$\uparrow$ & RMSE$\downarrow$ & Avg. Err.$\downarrow$ & Max Err.$\downarrow$ \\ \hline
w/o EC filter & 0.871 & 21.75 & 13.93 & 3.53 & 8.19 \\ \hline
w/ EC filter  & \textbf{0.889} & \textbf{22.39} & \textbf{13.38} & \textbf{3.38} & \textbf{5.18} \\ \hline
\end{tabular}
}
\end{table}

\textbf{DeArtifact Projection Optimization:} 
We propose the DeArtifact module to reduce artifacts that traditional projection compensation networks cannot avoid. A pre-trained ShadingNet serves as the projection network to provide WC-TPS with foundational training data. We further incorporate a real SL image pair into TPS training, adjusting data weights to improve photometric realism.

\begin{figure}[th]
  \centering
  \includegraphics[width=\linewidth]{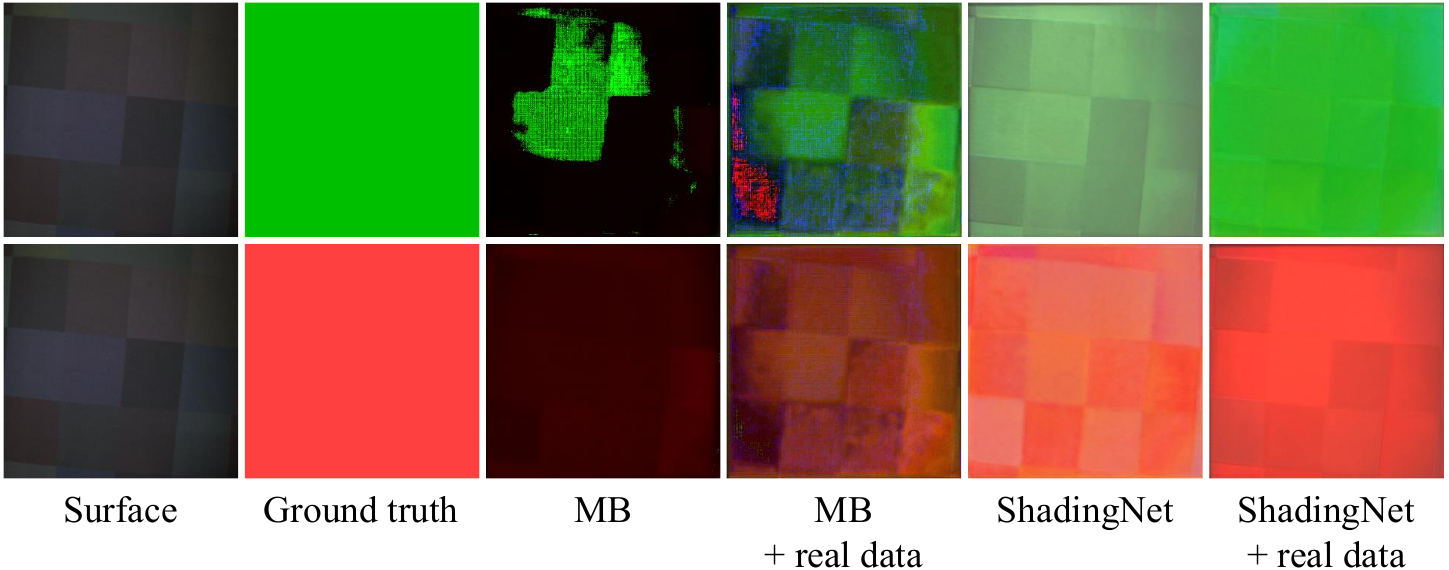}
  \caption{Contribution of different modules to photometric representation. ShadingNet offers a robust initial estimate, while real calibration data refines color-related biases.}
  \label{ablation_2}
\end{figure}

To evaluate the contributions of synthetic and real data, we conduct ablation experiments comparing the accuracy of resulting TPS photometric models. Without ShadingNet-generated data, we simulate projections using a multiplication blend (MB) mode processed by PCNet. As shown in Fig.~\ref{ablation_2}, this leads to significant color deviations or data loss. Adding real SL data improves results but still shows deviations. ShadingNet enables closer alignment with actual projection behavior, and combining it with real data allows fine-tuning toward the ground truth.

Our experiments are configured using the matched ground truth as input, thereby avoiding the influence introduced by Stage I. Table~\ref{table2.3} compares artifact reduction effectiveness via SSIM, PSNR, and RMSE. In DeArtifact, we first verify the TPS photometric mapping, then exclude pixels with unreliable information from optimization. Our experiments show three improvements: (1) smoother gradients in color transitions, (2) better spectral fidelity via optimized color coding, and (3) removal of grid discontinuities.

\begin{table}[h]
\centering
\caption{Contribution of different modules in Stage II to the photometric representation, while also demonstrating their optimization capability on the final decoding performance within binary coding at the highest encoding density.}
\setlength{\tabcolsep}{0.8mm}
{\fontsize{9}{10}\selectfont
\begin{tabular}{|c|c|c|c|c|c|c|}
\hline
WC-TPS & ShadingNet & Rea & SSIM$\uparrow$ & PSNR$\uparrow$ & RMSE$\downarrow$ & Err. (\%)$\downarrow$ \\ \hline
         &         &      & 0.760          & 19.99           & 24.62            & 6.31                  \\ \hline
\checkmark &       &      & 0.763          & 20.01           & 24.33            & 3.75                  \\ \hline
\checkmark & \checkmark &  & 0.823          & 21.77           & 22.36            & 1.22                  \\ \hline
\checkmark & \checkmark & \checkmark & \textbf{0.849} & \textbf{23.24} & \textbf{18.76} & \textbf{0.98}         \\ \hline
\end{tabular}
}
\label{table2.3}
\end{table}

\section{Conclusion}
This work addresses two key challenges in structured light: scene-specific calibration inefficiency and pattern-dependent optimization. We propose GUSLO, a general and unified framework that performs single-shot geometric calibration and artifact-aware photometric adaptation via physics-guided transfer. By integrating geometric interpolation with adaptive color mapping, GUSLO enables robust optimization across binary, speckle, and color-coded SL patterns without multi-pattern calibration. Experiments show consistent improvements under challenging lighting and surface textures, demonstrating its practicality for industrial inspection and cultural heritage digitization. Its compatibility with diverse encoding schemes offers a scalable foundation for future multi-device reconstruction systems.

\newpage
{

\begin{thebibliography}{34}
\providecommand{\natexlab}[1]{#1}
\providecommand{\url}[1]{\texttt{#1}}
\expandafter\ifx\csname urlstyle\endcsname\relax
  \providecommand{\doi}[1]{doi: #1}\else
  \providecommand{\doi}{doi: \begingroup \urlstyle{rm}\Url}\fi

\bibitem[Bak(2003)]{bak2003rapid}
David Bak.
\newblock Rapid prototyping or rapid production? 3d printing processes move industry towards the latter.
\newblock \emph{Assembly Automation}, 2003.

\bibitem[Community(2025)]{blender2025}
Blender~Online Community.
\newblock Blender – a 3d modeling and rendering package, 2025.

\bibitem[Ding and Liang(2024)]{ding2024digital}
Qian-Kun Ding and Hui-E Liang.
\newblock Digital restoration and reconstruction of heritage clothing: a review.
\newblock \emph{Heritage Science}, 12\penalty0 (1):\penalty0 225, 2024.

\bibitem[Dodda et~al.(2023)Dodda, Kuruguntla, Elumalai, Chinnadurai, Sheridan, and Muniraj]{dodda2023denoising}
Vineela~Chandra Dodda, Lakshmi Kuruguntla, Karthikeyan Elumalai, Sunil Chinnadurai, John~T Sheridan, and Inbarasan Muniraj.
\newblock A denoising framework for 3d and 2d imaging techniques based on photon detection statistics.
\newblock \emph{Scientific Reports}, 13\penalty0 (1):\penalty0 1365, 2023.

\bibitem[Dong et~al.(2023)Dong, Ling, and Huang]{dong2023adaptive}
Xin Dong, Haibin Ling, and Bingyao Huang.
\newblock Adaptive color structured light for calibration and shape reconstruction.
\newblock In \emph{IEEE International Symposium on Mixed and Augmented Reality (ISMAR)}, pages 1240--1249. IEEE, 2023.

\bibitem[Fu et~al.(2023)Fu, Fan, Jing, and Tan]{fu2023high}
Yichen Fu, Junfeng Fan, Fengshui Jing, and Min Tan.
\newblock High dynamic range structured light 3-d measurement based on region adaptive fringe brightness.
\newblock \emph{IEEE Transactions on Industrial Electronics}, 2023.

\bibitem[Fu et~al.(2024)Fu, Huang, Xiao, Li, Li, and Zuo]{fu2024deep}
Yanjun Fu, Yiliang Huang, Wei Xiao, Fangfang Li, Yunzhan Li, and Pengfei Zuo.
\newblock Deep learning-based binocular composite color fringe projection profilometry for fast 3d measurements.
\newblock \emph{Optics and Lasers in Engineering}, 172:\penalty0 107866, 2024.

\bibitem[Fujii et~al.(2005)Fujii, Grossberg, and Nayar]{fujii2005projector}
Kensaku Fujii, Michael~D Grossberg, and Shree~K Nayar.
\newblock A projector-camera system with real-time photometric adaptation for dynamic environments.
\newblock In \emph{IEEE Conference on Computer Vision and Pattern Recognition (CVPR)}, pages 814--821. IEEE, 2005.

\bibitem[Grundh{\"o}fer and Iwai(2015)]{grundhofer2015robust}
Anselm Grundh{\"o}fer and Daisuke Iwai.
\newblock Robust, error-tolerant photometric projector compensation.
\newblock \emph{IEEE Transactions on Image Processing}, 24\penalty0 (12):\penalty0 5086--5099, 2015.

\bibitem[Gupta(2007)]{gupta2007gray}
Pratibha Gupta.
\newblock \emph{Gray code composite pattern structured light illumination}.
\newblock PhD thesis, University of Kentucky Libraries, 2007.

\bibitem[Holl et~al.(2021)Holl, Pallas, and Bellendorf]{holl2021structured}
Kristina Holl, Leander Pallas, and Paul Bellendorf.
\newblock Structured light scanning as a monitoring method to investigate dimen-sional changes due to climatic changes on cultural heritage.
\newblock In \emph{Proceedings of the International Conference on Cultural Heritage and New Technologies}, 2021.

\bibitem[Huang and Ling(2019)]{huang2019end}
Bingyao Huang and Haibin Ling.
\newblock End-to-end projector photometric compensation.
\newblock In \emph{IEEE Conference on Computer Vision and Pattern Recognition (CVPR)}, pages 6810--6819, 2019.

\bibitem[Huang et~al.(2018)Huang, Ozdemir, Tang, Liao, and Ling]{huang2018single}
Bingyao Huang, Samed Ozdemir, Ying Tang, Chunyuan Liao, and Haibin Ling.
\newblock A single-shot-per-pose camera-projector calibration system for imperfect planar targets.
\newblock In \emph{2018 IEEE International Symposium on Mixed and Augmented Reality Adjunct (ISMAR-Adjunct)}, pages 15--20. IEEE, 2018.

\bibitem[Huang et~al.(2021)Huang, Sun, and Ling]{huang2021end}
Bingyao Huang, Tao Sun, and Haibin Ling.
\newblock End-to-end full projector compensation.
\newblock \emph{IEEE Transactions on Pattern Analysis and Machine Intelligence}, 44\penalty0 (6):\penalty0 2953--2967, 2021.

\bibitem[Jia et~al.(2024)Jia, Li, Yang, Lin, Liu, and Chen]{jia2024adaptivestereo}
Tong Jia, Xiaofang Li, Xiao Yang, Shuyang Lin, Yizhe Liu, and Dongyue Chen.
\newblock Adaptivestereo: Depth estimation from adaptive structured light.
\newblock \emph{Optics \& Laser Technology}, 169:\penalty0 110076, 2024.

\bibitem[Kawasaki et~al.(2008)Kawasaki, Furukawa, Sagawa, and Yagi]{kawasaki2008dynamic}
Hiroshi Kawasaki, Ryo Furukawa, Ryusuke Sagawa, and Yasushi Yagi.
\newblock Dynamic scene shape reconstruction using a single structured light pattern.
\newblock In \emph{IEEE Conference on Computer Vision and Pattern Recognition (CVPR)}, pages 1--8. IEEE, 2008.

\bibitem[Koninckx et~al.(2005)Koninckx, Peers, Dutr{\'e}, and Van~Gool]{koninckx2005scene}
Thomas~P Koninckx, Pieter Peers, Philip Dutr{\'e}, and Luc Van~Gool.
\newblock Scene-adapted structured light.
\newblock In \emph{IEEE Conference on Computer Vision and Pattern Recognition (CVPR)}, pages 611--618. IEEE, 2005.

\bibitem[Li et~al.(2022)Li, Qian, Feng, Chen, and Zuo]{li2022deep}
Yixuan Li, Jiaming Qian, Shijie Feng, Qian Chen, and Chao Zuo.
\newblock Deep-learning-enabled dual-frequency composite fringe projection profilometry for single-shot absolute 3d shape measurement.
\newblock \emph{Opto-Electronic Advances}, 5\penalty0 (5):\penalty0 210021--1, 2022.

\bibitem[Li et~al.(2023)Li, Yin, Li, and Xie]{li2023physics}
Yuqi Li, Wenting Yin, Jiabao Li, and Xijiong Xie.
\newblock Physics-based efficient full projector compensation using only natural images.
\newblock \emph{IEEE Transactions on Visualization and Computer Graphics}, 2023.

\bibitem[Montusiewicz et~al.(2021)Montusiewicz, Mi{\l}osz, K{\k{e}}sik, and {\.Z}y{\l}a]{montusiewicz2021structured}
Jerzy Montusiewicz, Marek Mi{\l}osz, Jacek K{\k{e}}sik, and Kamil {\.Z}y{\l}a.
\newblock Structured-light 3d scanning of exhibited historical clothing—a first-ever methodical trial and its results.
\newblock \emph{Heritage Science}, 9\penalty0 (1):\penalty0 74, 2021.

\bibitem[Nguyen et~al.(2022)Nguyen, Ly, Qiong~Li, and Wang]{nguyen2022single}
Andrew-Hieu Nguyen, Khanh~L Ly, Charlotte Qiong~Li, and Zhaoyang Wang.
\newblock Single-shot 3d shape acquisition using a learning-based structured-light technique.
\newblock \emph{Applied Optics}, 61\penalty0 (29):\penalty0 8589--8599, 2022.

\bibitem[Pages et~al.(2003)Pages, Salvi, Garcia, and Matabosch]{1241585}
J. Pages, J. Salvi, R. Garcia, and C. Matabosch.
\newblock Overview of coded light projection techniques for automatic 3d profiling.
\newblock In \emph{IEEE International Conference on Robotics and Automation}, pages 133--138 vol.1, 2003.

\bibitem[Park et~al.(2022)Park, Jung, and Moon]{park2022projector}
Jino Park, Donghyuk Jung, and Bochang Moon.
\newblock Projector compensation framework using differentiable rendering.
\newblock \emph{IEEE Access}, 10:\penalty0 44461--44470, 2022.

\bibitem[Qian et~al.(2019)Qian, Feng, Tao, Hu, Liu, Wu, Chen, and Zuo]{qian2019high}
Jiaming Qian, Shijie Feng, Tianyang Tao, Yan Hu, Kai Liu, Shuaijie Wu, Qian Chen, and Chao Zuo.
\newblock High-resolution real-time 360 3d model reconstruction of a handheld object with fringe projection profilometry.
\newblock \emph{Optics Letters}, 44\penalty0 (23):\penalty0 5751--5754, 2019.

\bibitem[Qian et~al.(2021)Qian, Feng, Xu, Tao, Shang, Chen, and Zuo]{qian2021high}
Jiaming Qian, Shijie Feng, Mingzhu Xu, Tianyang Tao, Yuhao Shang, Qian Chen, and Chao Zuo.
\newblock High-resolution real-time 360° 3d surface defect inspection with fringe projection profilometry.
\newblock \emph{Optics and Lasers in Engineering}, 137:\penalty0 106382, 2021.

\bibitem[Raskar et~al.(2002)Raskar, van Baar, and Chai]{raskar2002low}
Ramesh Raskar, Jeroen van Baar, and Jin~Xiang Chai.
\newblock A low-cost projector mosaic with fast registration.
\newblock In \emph{Asian Conference on Computer Vision (ACCV)}, 2002.

\bibitem[Sansoni et~al.(2009)Sansoni, Trebeschi, and Docchio]{sansoni2009state}
Giovanna Sansoni, Marco Trebeschi, and Franco Docchio.
\newblock State-of-the-art and applications of 3d imaging sensors in industry, cultural heritage, medicine, and criminal investigation.
\newblock \emph{Sensors}, 9\penalty0 (1):\penalty0 568--601, 2009.

\bibitem[Shih et~al.(2020)Shih, Liu, Shyu, and Chen]{shih2020enhancement}
Kuang-Tsu Shih, Jen-Shuo Liu, Frank Shyu, and Homer~H Chen.
\newblock Enhancement and speedup of photometric compensation for projectors by reducing inter-pixel coupling and calibration patterns.
\newblock \emph{IEEE Transactions on Image Processing}, 30:\penalty0 418--430, 2020.

\bibitem[Stelzner et~al.(2022)Stelzner, Stelzner, Martinez-Garcia, Gwerder, Wittk{\"o}pper, Muskalla, Cramer, Heinz, Egg, and Schuetz]{stelzner2022stabilisation}
J{\"o}rg Stelzner, Ingrid Stelzner, Jorge Martinez-Garcia, Damian Gwerder, Markus Wittk{\"o}pper, Waldemar Muskalla, Anja Cramer, Guido Heinz, Markus Egg, and Philipp Schuetz.
\newblock Stabilisation of waterlogged archaeological wood: the application of structured-light 3d scanning and micro computed tomography for analysing dimensional changes.
\newblock \emph{Heritage Science}, 10\penalty0 (1):\penalty0 60, 2022.

\bibitem[Sugimoto et~al.(2021)Sugimoto, Iwai, Ishida, Punpongsanon, and Sato]{sugimoto2021directionally}
Masatoki Sugimoto, Daisuke Iwai, Koki Ishida, Parinya Punpongsanon, and Kosuke Sato.
\newblock Directionally decomposing structured light for projector calibration.
\newblock \emph{IEEE Transactions on Visualization and Computer Graphics}, 27\penalty0 (11):\penalty0 4161--4170, 2021.

\bibitem[Wang et~al.(2024)Wang, Song, Wang, Ren, Zhao, Dou, Di, Barbastathis, Zhou, Zhao, et~al.]{wang2024use}
Kaiqiang Wang, Li Song, Chutian Wang, Zhenbo Ren, Guangyuan Zhao, Jiazhen Dou, Jianglei Di, George Barbastathis, Renjie Zhou, Jianlin Zhao, et~al.
\newblock On the use of deep learning for phase recovery.
\newblock \emph{Light: Science \& Applications}, 13\penalty0 (1):\penalty0 4, 2024.

\bibitem[Xu et~al.(2023)Xu, Lin, Zhou, Zeng, Yu, Zhou, and Wu]{xu2023unified}
Xianmin Xu, Yuxin Lin, Haoyang Zhou, Chong Zeng, Yaxin Yu, Kun Zhou, and Hongzhi Wu.
\newblock A unified spatial-angular structured light for single-view acquisition of shape and reflectance.
\newblock In \emph{IEEE Conference on Computer Vision and Pattern Recognition (CVPR)}, pages 206--215, 2023.

\bibitem[Yamada et~al.(2024)Yamada, Takatori, Adachi, Hasegawa, Li, Pan, Thufail, Yamaguchi, and Tanaka]{yamada2024high}
Yuri Yamada, Satoshi Takatori, Motoaki Adachi, Kyoko Hasegawa, Liang Li, Jiao Pan, Fadjar~I Thufail, Hiroshi Yamaguchi, and Satoshi Tanaka.
\newblock High-visibility edge-highlighting visualization of 3d scanned point clouds based on dual 3d edge extraction.
\newblock \emph{Remote Sensing}, 16\penalty0 (19):\penalty0 3750, 2024.

\bibitem[Zou et~al.(2024)Zou, Yu, Huang, and Zhao]{Zou_2024_CVPR}
Zhen Zou, Wei Yu, Jie Huang, and Feng Zhao.
\newblock Semantic pre-supplement for exposure correction.
\newblock In \emph{IEEE Conference on Computer Vision and Pattern Recognition Workshops (CVPRW)}, pages 5961--5970, 2024.

\end{thebibliography}

}

\clearpage
\setcounter{page}{1}
\maketitlesupplementary

\section{Construction of the Real Dataset}
\begin{figure}[th]
  \centering
  \includegraphics[width=\linewidth]{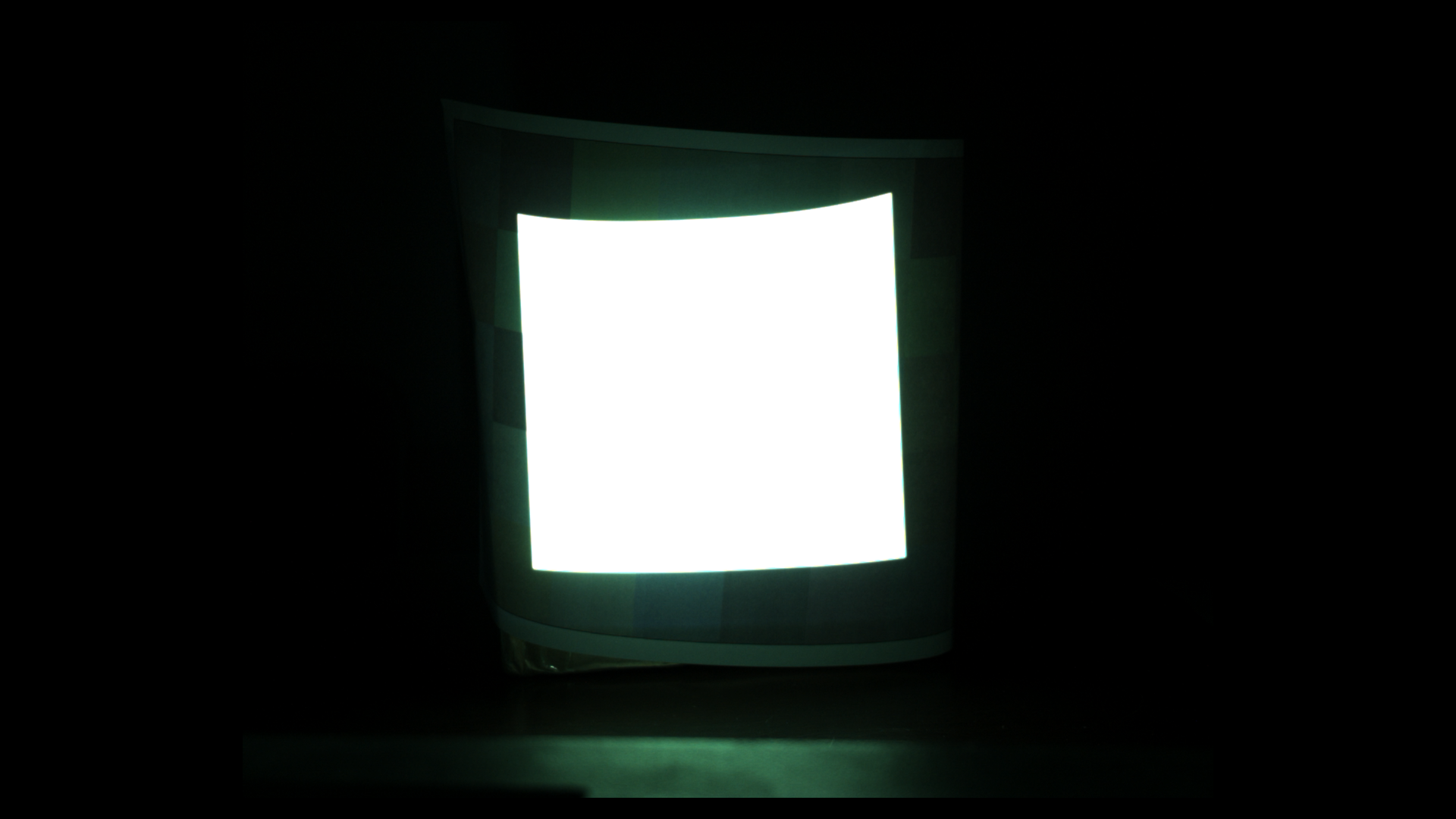}
  \caption{The effective projection area is obtained by increasing the amount of incoming light from the projector.}
  \label{ROI}
\end{figure}

The real dataset needs to satisfy the training requirements of global matching and photometric compensation. In this experiment, we utilize two complementary sets of 10-bit vertical and horizontal Gray codes to construct the matching dataset. To reasonably filter out mismatched codes and enhance dataset quality, we increase the camera exposure time to amplify the projector-originated light captured by the camera. During this process, the projector projects a pure white pattern to effectively acquire its valid projection range, as shown in Fig. \ref{ROI}. We create a mask based on this image and apply it to the previously captured vertical and horizontal complementary Gray code sets. Mismatched codes typically correspond to regions where the mask is 0 (displayed as pure black in the image). This allows us to effectively eliminate most mismatched codes caused by Gray code decoding.
Since our proposed triangulation interpolation algorithm is applicable to most discrete-to-continuous matching algorithms, after obtaining the discrete Gray code matching with error codes removed, we use our proposed triangulation interpolation algorithm to convert it into continuous global matching. The mapping relationship of each pixel is then saved, laying the foundation for rapidly generating training data for projection compensation in the next stage.

To enhance the robustness of the photometric compensation network, we prioritize color-diverse projection patterns by implementing an entropy-based screening mechanism. The workflow initiates with RGB histogram computation and progresses through entropy evaluation as follows:

First, the RGB color space is discretized into $k^3$ uniformly distributed bins. Let $I\in \mathbb{R}^{w\times h\times 3}$ denote the captured image with width w and height h. We compute the 3D histogram $hist\in \mathbb{N}^{k\times k\times k}$ by accumulating pixel counts within each chromaticity bin:

\begin{equation}
\label{eqa1}
\text{hist}[i,j,k] = \sum_{x=1}^{w}\sum_{y=1}^{h} \delta(I(x,y), (i,j,k))
\end{equation}
where $\delta \left( \cdot \right)$ is the Kronecker delta function indicating bin membership.

The histogram is subsequently normalized to obtain probability estimates:
\begin{equation}
\label{eqa2}
p_{i,j,k} = \frac{\text{hist}[i,j,k]}{\sum_{m=1}^{k}\sum_{n=1}^{k}\sum_{o=1}^{k} \text{hist}[m,n,o]}
\end{equation}

We then quantify color richness through Shannon entropy:
\begin{equation}
\label{eqa3}
H = -\sum_{i=1}^{k}\sum_{j=1}^{k}\sum_{k=1}^{k} p_{i,j,k} \log_2(p_{i,j,k} + \epsilon)
\end{equation}
where $\epsilon =10^{-6}$  prevents numerical instability. Images satisfying $ H \geq T $ are retained as candidate patterns, while low-entropy samples are discarded to avoid chromatic monotony in training data. Generally, $T$ is set to 6.

\section{Construction of the Synthetic Dataset}
We set up the Open Structured Light Benchmarking System in Blender. First, the projection effect is achieved using the Projector plugin (add-on available from https://github.com/Ocupe/Projectors). To make the camera more closely resemble a real-world model, we incorporate the Photographer5 plugin (add-on available from https://chafouin.gumroad.com/l/photographer5) to adjust exposure and lens parameters, as well as the Per-Camera Resolution plugin (add-on available from https://extensions.blender.org/add-ons/per-camera-resolution/) to modify the camera's resolution.

Since the method of calculating camera poses in Blender differs from the coordinate system setup in structured light systems, we developed a coordinate transformation script. This script ensures that users can correctly export the poses of all sensors within the environment.

To facilitate the automated generation of datasets, we have predefined several scene layouts and set the material, texture maps, position of the test object, and the intensity and position of ambient lighting as configurable parameters stored in a configuration file. This allows for the automatic generation of a shooting task sequence by merely adjusting the configuration file, enabling users to produce their own synthetic datasets.

\section{Supplementary Results}
\subsection{Comparison of Global Matching Methods}
\begin{figure}[ht]
  \centering  \includegraphics[width=\linewidth]{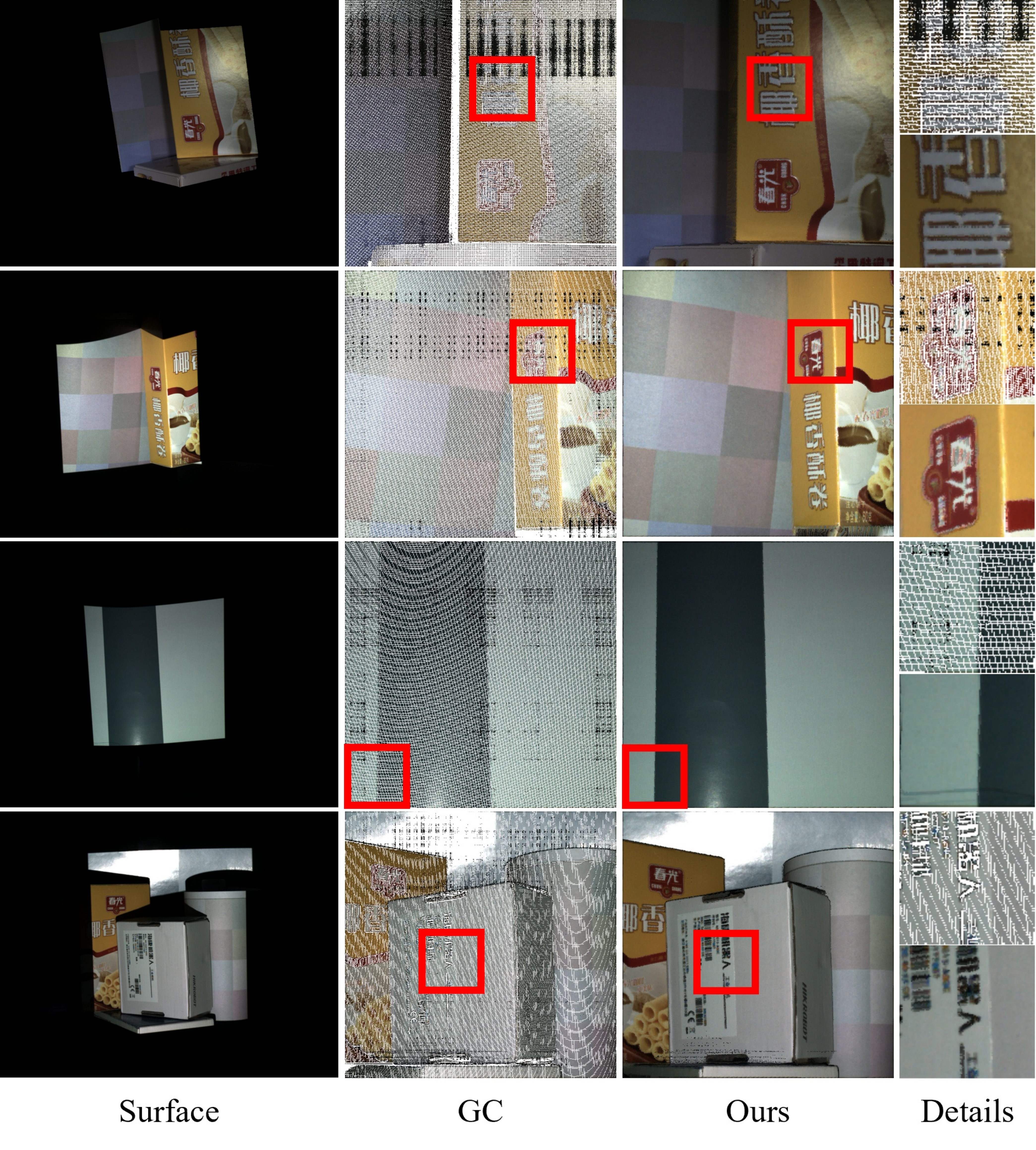}
  \caption{Comparison with the method of global encoding using Gray code (GC).}
  \label{app_3}
\end{figure}

In the comparison of global matching experiments, we contrasted 42 Gray-code patterns with the WarpingNet method utilizing 400 images. However, during the actual projection process, periodic decoding errors occur in the Gray-code patterns, manifesting as black regions in the matching results, as illustrated in Fig. \ref{app_3}. These erroneous regions were filtered out before proceeding with subsequent decoding optimization comparisons against our method.

\begin{figure}[ht]
  \centering  \includegraphics[width=\linewidth]{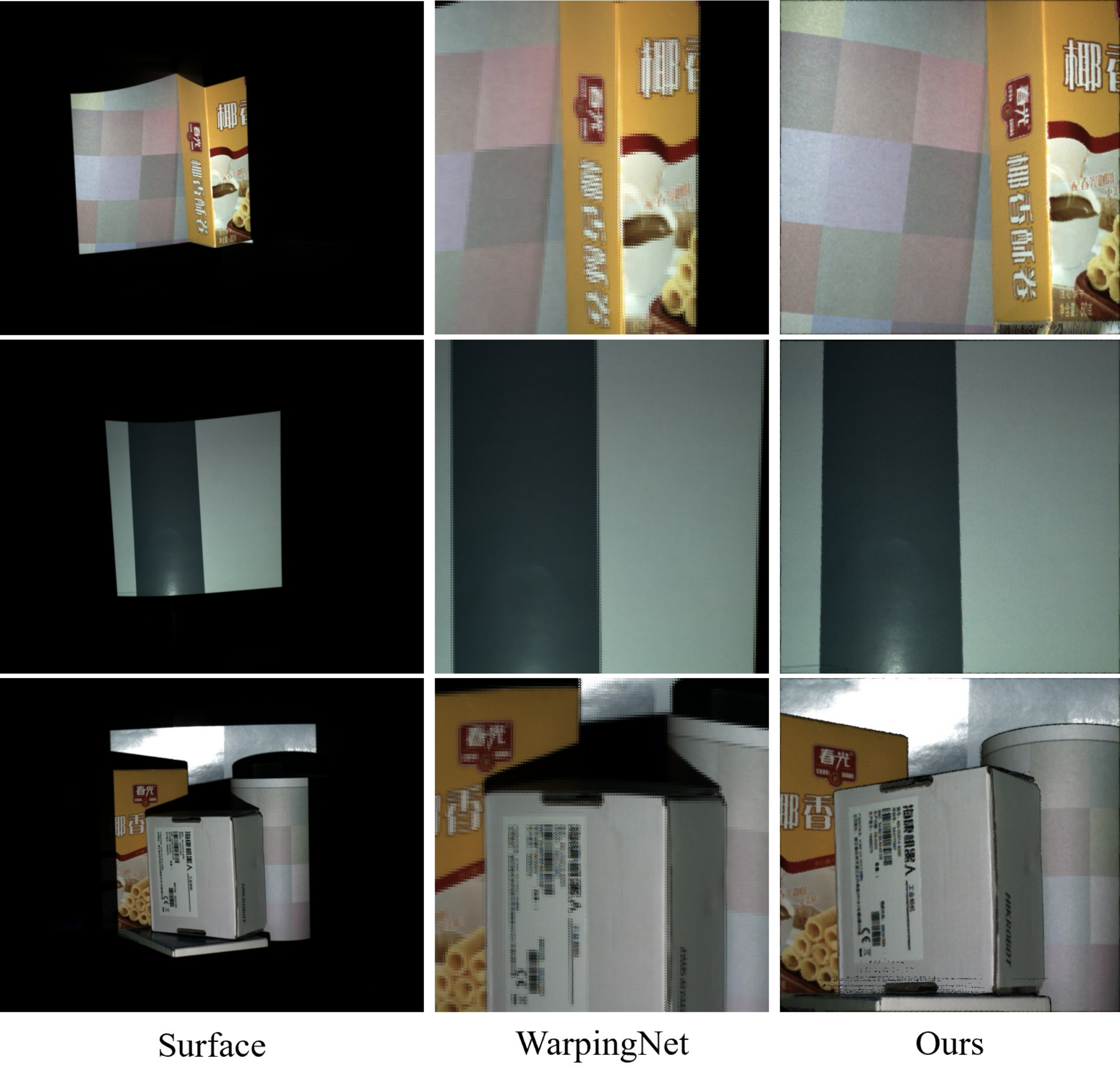}
  \caption{Comparison with the method of global encoding using WarpingNet.}
  \label{app_4}
\end{figure}

Although WarpingNet can achieve effective global matching on single smooth surfaces, deep learning-based approaches generally exhibit inferior robustness across varying illumination conditions compared to traditional encoding-based matching schemes (such as Gray Code and our GMM). This limitation poses certain risks in structured-light 3D reconstruction, an application demanding high precision. As shown in the second row of Fig. \ref{app_4}, WarpingNet fails to properly unwrap the single surface, resulting in misaligned black borders along the image periphery.

\subsection{Implementation details and geometric robustness}
 PCNet’s siamese encoder is trained using combined L1 and perceptual losses on both surface and projection domains; illumination–reflectance separation is achieved through feature subtraction between shared-weight encoders.
 
 WarpingNet fails on discontinuous surfaces because its convolutional smoothness prior enforces spatial coherence, causing cross-surface blending. GUSLO, in contrast, uses Delaunay-based piecewise interpolation that preserves local boundaries and avoids this issue.
 
 Our barycentric interpolation (Eq. 4–8) also handles occlusions and non-convex surfaces robustly. Tests on synthetic surfaces with varying sharpness (amp/sig) show:
 
\begin{tabular}{|l|c|c|c|c|}
\hline
amp/sig & 3.0/0.3 & 3.0/0.7 & 5.0/0.3 & 5.0/0.7 \\
\hline
GC & 0.362 & 0.357 & 0.355 & 0.354 \\
GC+Linear & 0.927 & 0.925 & 0.899 & 0.895 \\
GC+GMM & 0.952 & 0.943 & 0.916 & 0.909 \\
GMM & 0.889 & 0.877 & 0.877 & 0.851 \\
\hline
\end{tabular}

Even in the sharpest case (amp/sig=3.0/0.3), GMM differs by only 2.12\% from GC+GMM (3.38\% vs 3.31\%), confirming stable single-shot calibration.

\subsection{Optimization Effects of Different Structured Light Patterns}
Since our structured light system has not undergone intrinsic and extrinsic calibration of the camera and projector, and given that our algorithm is designed for convenient use without reliance on such calibration parameters, we are unable to reconstruct the absolute 3D geometry of the target object. However, we can evaluate and compare the optimization performance of different methods using decoding error rates. Correct decoding implies that depth information has been successfully computed for the corresponding pixel. Therefore, the visualized decoding results implicitly reflect the geometric accuracy of the object in 3D reconstruction.

\subsubsection{Binary Stripe Structured Light Pattern}
\begin{figure*}[h]
  \centering  \includegraphics[width=\linewidth]{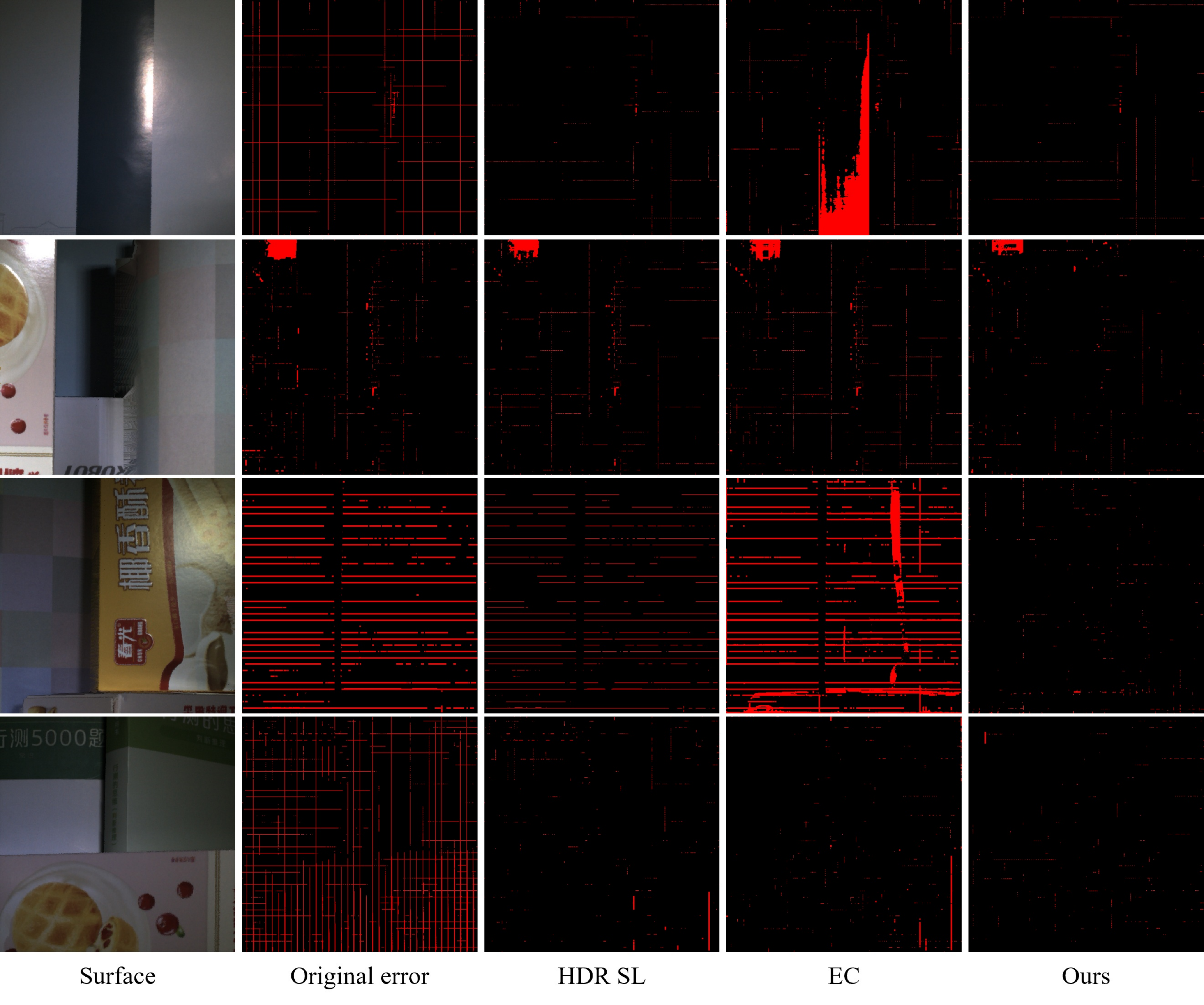}
  \caption{The optimization results of different methods for binary stripe structured light are shown in the figure, where the red sections denote the areas with decoding errors.}
  \label{app_1}
\end{figure*}

During real world projection process, Gray code patterns are most susceptible to decoding errors at the transition edges between code values. Additionally, underexposure or overexposure caused by object surface materials and lighting conditions can reduce the contrast of binary fringes, thereby degrading decoding accuracy. Figure \ref{app_1} presents several results from real-world scenarios, demonstrating that the optimization performance of our algorithm is particularly evident and more pronounced in practical applications.

\subsubsection{Speckle Structured Light Pattern}
\begin{figure*}[h]
  \centering  \includegraphics[width=\linewidth]{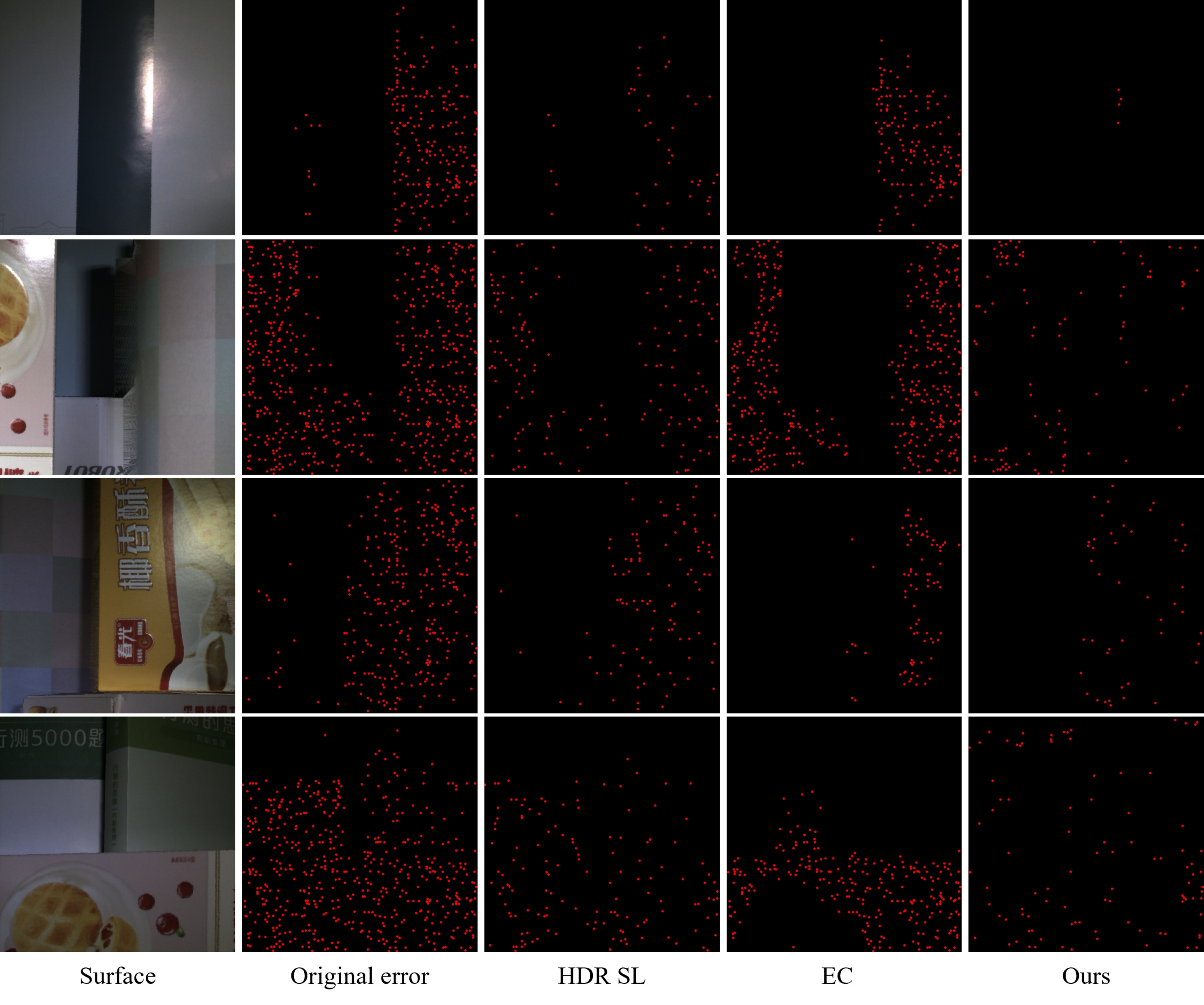}
  \caption{The optimization results of different methods for speckle structured light are shown in the figure, where the red sections denote the areas with decoding errors.}
  \label{app_2}
\end{figure*}

\begin{figure*}[ht]
  \centering  \includegraphics[width=\linewidth]{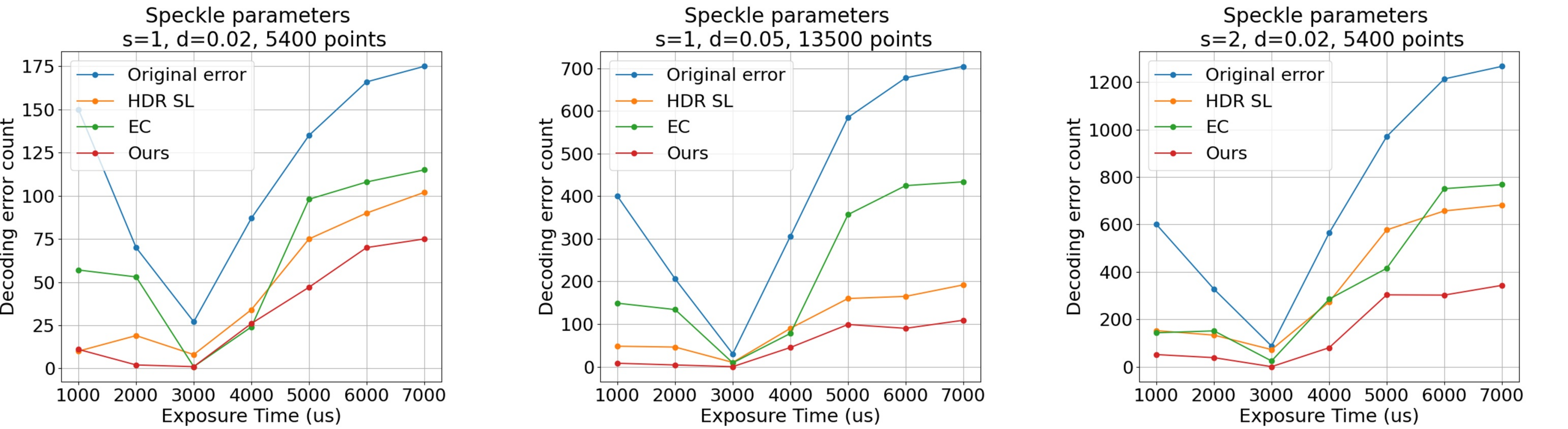}
  \caption{The optimization results of different methods for speckle structured light are shown in the figure, where the red sections denote the areas with decoding errors.}
  \label{app_7}
\end{figure*}

Since speckle-based structured light relies on discrete matching, and speckles possess distinct characteristics, the fundamental attributes being density (d) and spot size (s), we additionally evaluated the optimization performance of different algorithms under varying exposure times. Exposure time is a camera parameter independent of ambient illumination, object properties, and the type of projected structured light; therefore, the experimental results concerning exposure are generally applicable across different types of structured light. We tested three distinct speckle patterns and conducted stepwise evaluations across a common exposure range from 1000\textmu s to 7000\textmu s. As shown in Fig. \ref{app_7}, under the current projector illumination level, the optimal exposure time is approximately 3000\textmu s. Our method consistently achieves the best optimization performance across all tested exposure settings.

\subsubsection{Colored Structured Light Pattern}
\begin{figure*}[h]
  \centering  \includegraphics[width=\linewidth]{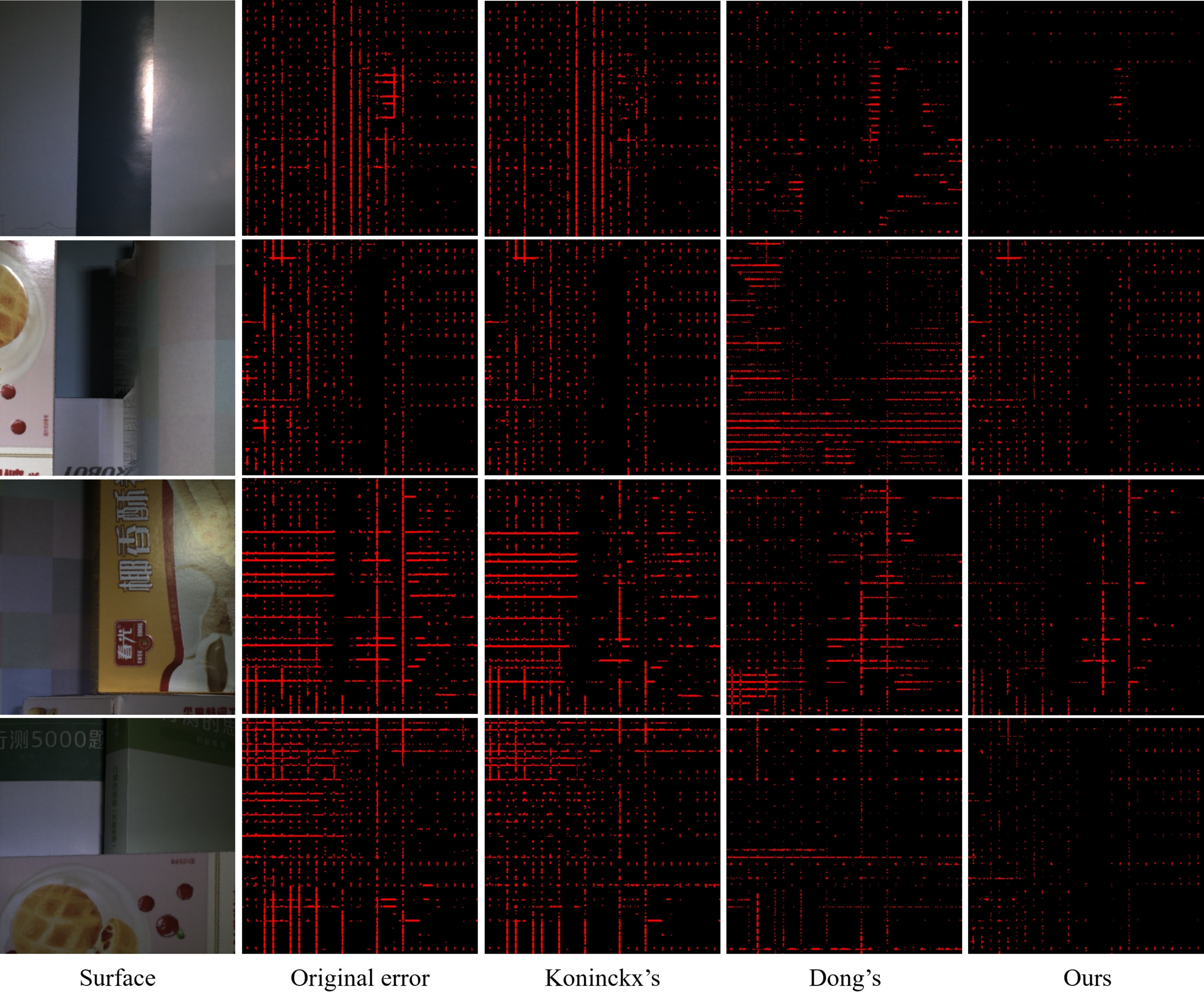}
  \caption{The optimization results of different methods for colored structured light are shown in the figure, where the red sections denote the areas with decoding errors.}
  \label{app_6}
\end{figure*}

Colored structured light, possessing three-dimensional color information, enables single-projection matching. However, this also makes its decoding accuracy highly dependent on precise color recognition, which generally results in lower decoding precision compared to binary stripe and speckle-based structured light. Figure \ref{app_6} shows the experimental results in real-world scenarios. Given that our algorithm can perform corresponding color compensation based on the texture of the object's surface, it exhibits a significant advantage in optimizing the performance of colored structured light.

\begin{figure*}[h]
  \centering  \includegraphics[width=\linewidth]{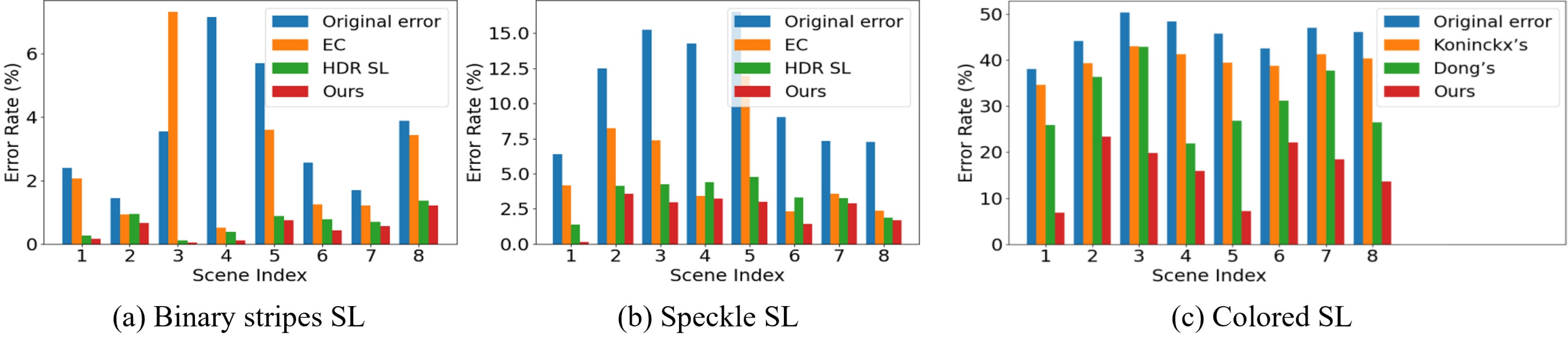}
  \caption{A comparative evaluation of different optimization methods for three common structured light techniques across various real-world scenarios.}
  \label{app_5}
\end{figure*}

We conducted several sets of experiments by varying the lighting and projection angles in these scenarios. The experimental results are shown in Figure \ref{app_5}.

\subsection{Runtime and hardware diversity}
On an RTX 3090 with 1024×1024 inputs, geometric calibration requires 164 ms, photometric calibration (online training) 119.78 s, and per-frame optimization 1.37 s during inference. Since optimization can be performed prior to 3D reconstruction, projection patterns can be pre-optimized to ensure higher reconstruction accuracy. Additionally, we constructed datasets in simulation with varied camera and projector parameters, providing broader hardware diversity beyond the physical setup.

\section{Ablation Study}
\begin{figure*}[h]
  \centering  \includegraphics[width=\linewidth]{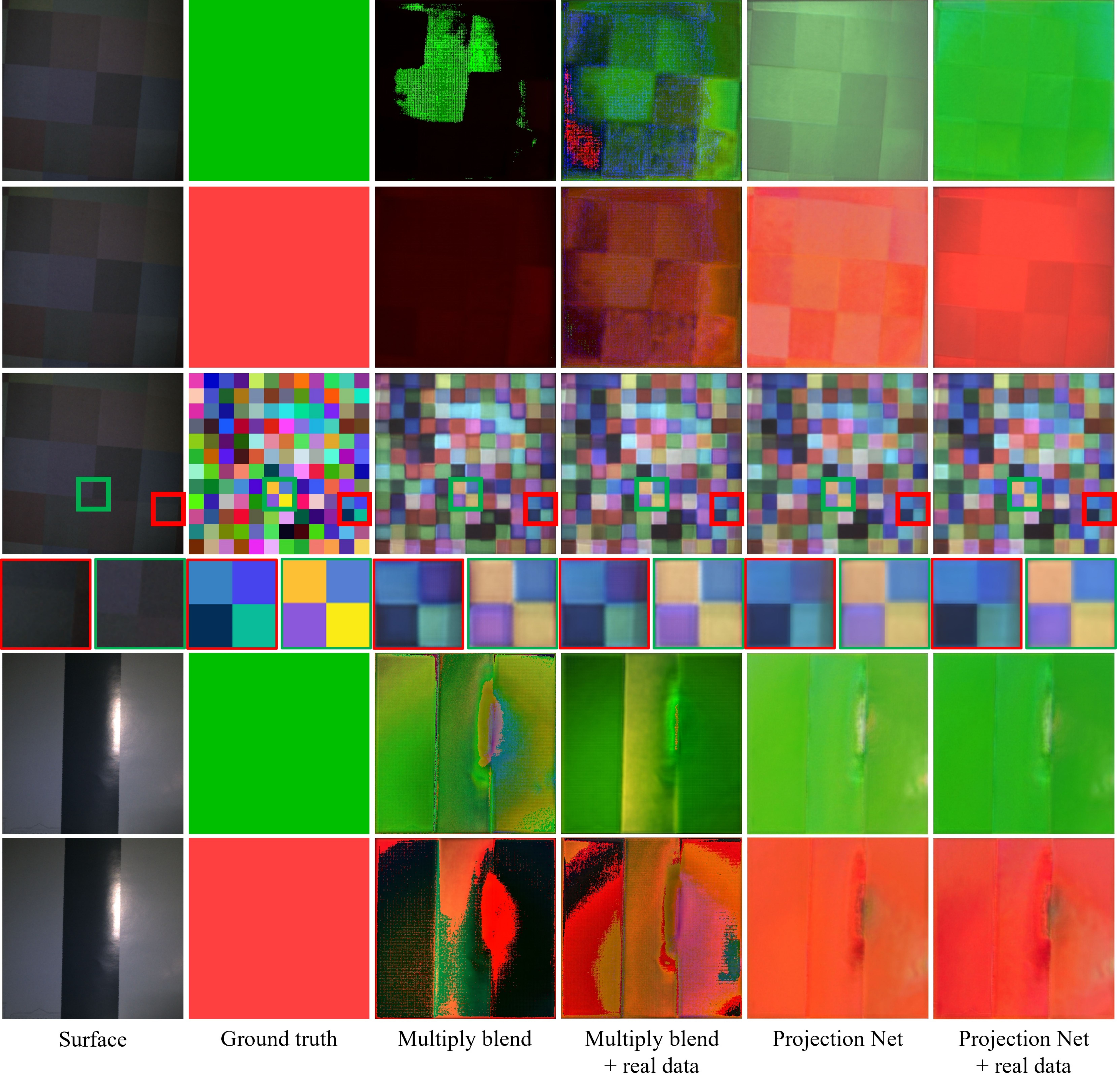}
  \caption{The photometric results of the WC-TPS function.}
  \label{app_8}
\end{figure*}

Figure \ref{app_8} showcases the fitting degree of each module within the DeArtifact module to the photometric representation. It is evident from the comparison of solid color images that all modules in Stage II play a crucial role in enabling the system to accurately understand the photometric expressions.

\end{document}